\documentclass[journal]{IEEEtran}
\ifCLASSINFOpdf
\else
\fi
\usepackage{amsmath,amsfonts}
\usepackage{algorithmic}
\usepackage{algorithm}
\usepackage{array}
\usepackage[caption=false,font=normalsize,labelfont=sf,textfont=sf]{subfig}
\usepackage{textcomp}
\usepackage{stfloats}
\usepackage{url}
\usepackage{verbatim}
\usepackage{graphicx}
\usepackage{cite}
\usepackage{booktabs, multirow} 
\usepackage{color}

\usepackage{booktabs}
\usepackage{multirow}
\usepackage{adjustbox}
\usepackage{graphicx}

\usepackage{tabularx}
\usepackage{makecell}
\usepackage{threeparttable}

\begin{document}
%
\title{Cross-View Feature Matching: Survey, Benchmarking, and Foundation-Model Perspectives}

%
%
%

\author{Songlin~Du,~\IEEEmembership{Senior Member,~IEEE,}
        Xiaoyong~Lu,
        Zeyu~Wu,
        Xiaobo~Lu,
        Guobao~Xiao,~\IEEEmembership{Senior Member,~IEEE,}
        Bin~Fan,~\IEEEmembership{Senior Member,~IEEE,}
        Jiayi~Ma,~\IEEEmembership{Senior Member,~IEEE,}
        and~Takeshi~Ikenaga,~\IEEEmembership{Senior Member,~IEEE}
\thanks{This work was supported in part by the National Natural Science Foundation of China under grant 62472312, in part by the Shenzhen Science and Technology Program under grant JCYJ20240813161801002, and in part by the Guangdong Basic and Applied Basic Research Foundation under grants 2025A1515011943 and 2026A1515012748. \emph{(Corresponding author: Songlin Du.)}}
\thanks{Songlin~Du, Xiaoyong~Lu, Zeyu~Wu and Xiaobo~Lu are with the School of Automation, Southeast University, Nanjing 210096, China (e-mail: sdu@seu.edu.cn; luxiaoyong@seu.edu.cn; 220242094@seu.edu.cn; xblu@seu.edu.cn).}
\thanks{Guobao~Xiao is with the School of Computer Science and Technology, Tongji University, Shanghai 201804, China (e-mail: gbx@tongji.edu.cn).}
\thanks{Bin~Fan is with the School of Intelligence Science and Technology and the Institute of Artificial Intelligence, University of Science and Technology Beijing, Beijing 100083, China (e-mail: bin.fan@ieee.org).}
\thanks{Jiayi~Ma is with the Electronic Information School and the School of Robotics, Wuhan University, Wuhan 430072, China (e-mail: jyma2010@gmail.com).}
\thanks{Takeshi~Ikenaga is with the Graduate School of Information, Production and Systems, Waseda University, Kitakyushu 808-0135, Japan (e-mail: ikenaga@waseda.jp).}
}

\maketitle

\begin{abstract}
Cross-view feature matching aims to establish reliable correspondences across images with large viewpoint variations. Over the past decade, the field has evolved from task-specific models toward increasingly unified and generalizable correspondence models, with recent progress further driven by the emergence of vision foundation models (VFMs). Despite these advances, existing studies remain highly diverse in their problem formulations, model architectures, training paradigms, and evaluation protocols, making it difficult to obtain a unified understanding of the field. In this survey, we present a unified review of cross-view feature matching. We first introduce a structured taxonomy covering feature extraction, single-type feature matcher, multi-type feature matcher, VFMs based methods, training strategy and robust estimation, providing a coherent framework for analysis and comparison. We further examine recent advances, distilling key design principles and highlighting the shift toward unified and generalizable correspondence models. We also provide a unified experimental benchmarking of representative state-of-the-art methods under consistent protocols, enabling fair and comprehensive performance comparisons. In addition, we discuss open challenges and future directions, including efficiency, robustness under extreme conditions, and cross-domain generalization. This survey aims to provide a comprehensive and structured reference for understanding the evolution, current landscape, and future development of cross-view feature matching in the era of vision foundation models.
\end{abstract}


%
\IEEEpeerreviewmaketitle

\section{Introduction}
\IEEEPARstart{C}{ross-view} feature matching constitutes a fundamental problem in computer vision, concerned with establishing reliable correspondences between visual observations acquired from distinct viewpoints. Its core significance lies in enabling machines to bridge geometric, photometric, and perspective-induced discrepancies, thereby forming a coherent understanding of a shared scene despite substantial appearance variations. This capability is indispensable for constructing consistent representations of the physical world, particularly in scenarios where viewpoint changes introduce challenges such as nonlinear distortions, occlusions, and scale variations. The broader importance of cross-view feature matching extends to its role as an enabling mechanism for high-level visual understanding and spatial reasoning. By establishing reliable correspondences across disparate viewpoints, it supports the alignment and integration of multi-view observations, facilitating consistent scene interpretation and coherent geometric reasoning. This capability is fundamental to a range of core vision tasks, such as structure-from-motion (SfM) \cite{SfM}, visual localization \cite{Localization}, and simultaneous localization and mapping (SLAM) \cite{SLAM}. Consequently, cross-view feature matching serves as a critical building block for advancing robust perception systems capable of operating reliably in diverse, large-scale, and unconstrained real-world environments.

In recent years, cross-view feature matching has advanced rapidly, evolving from handcrafted pipelines to deep learning-based end-to-end frameworks with improved robustness and generalization. Since SuperGlue \cite{superglue} established the first successful attention-based matching framework in 2020, cross-view feature matching has experienced an unprecedented surge of innovation. Within only a few years, the field has progressed from graph-based sparse matching \cite{superglue} to Transformer-driven dense correspondence estimation \cite{loftr}, Mamba-based correspondence framework \cite{jamma,MambaGlue}, efficient adaptive matching architectures \cite{lightglue}, and, more recently, vision foundation models (VFMs) enhanced representations \cite{OmniGlue,roma,leroy2024grounding,wang2025vggt}. These advances have substantially reshaped the problem formulation, feature representation, matching strategy, and training paradigm of modern correspondence estimation. However, the rapid growth of the literature has also led to increasing fragmentation, with newly proposed methods differing significantly in their assumptions, architectures, and evaluation protocols. As a result, researchers and practitioners lack a unified view of the methodological landscape and the key factors driving recent progress. This motivates the need for a comprehensive and up-to-date survey that systematically reviews, categorizes, and analyzes contemporary cross-view feature matching approaches.

Unlike existing surveys that mainly trace the evolution from handcrafted pipelines to deep learning-based methods \cite{ma2021image,huang2024survey}, or focus on multimodal image matching and its applications \cite{jiang2021review}, our work specifically treats cross-view feature matching as a unified problem of geometric correspondence. Rather than organizing methods solely by algorithmic families or data modalities, we introduce a taxonomy grounded in correspondence modeling principles under cross-view settings, with particular emphasis on modern learning paradigms such as Transformer-, GNN-, Mamba-, and diffusion-based matchers. We further integrate recent vision foundation models into the discussion and analyze their interaction with task-specific matching frameworks, revealing both opportunities and gaps in current hybrid systems. Complementing the taxonomy, we provide a unified benchmarking and empirical evaluation under consistent protocols, enabling fair comparisons and offering practical insights into the strengths and limitations of representative methods. Key open challenges are also examined, including uncertainty-aware correspondence, explicit geometric reasoning, and scalable global matching. The main contributions of this survey are summarized as follow:
\begin{itemize}
\item \textbf{Hierarchical taxonomy.}
We propose a structured and hierarchical taxonomy that organizes existing methods along multiple orthogonal dimensions, including {feature extraction}, {single-type feature matcher}, {multi-type feature matcher}, {VFMs based methods}, {training strategy} and {robust estimation}, enabling fine-grained comparison across diverse design choices.
\item \textbf{Methodological analysis.}
We review recent advances in cross-view feature matching, including GNN-, Transformer-, Mamba-, and diffusion-based methods. We distill key design principles such as global context modeling, cross-image interaction, multi-scale reasoning, and multimodal fusion. We also summarize training strategies and robust estimation techniques that integrate self-supervision and geometric consistency for improved matching and outlier handling.
\item \textbf{Unified benchmarking.}
We establish a unified benchmarking protocol by evaluating representative methods on multiple standard datasets, \textit{e.g.}, MegaDepth, ScanNet and HPatches, under consistent metrics and settings, providing fair and comprehensive performance comparisons.
\item \textbf{Challenges and trends.}
We identify key challenges in cross-view feature matching, including the trade-off between geometric precision and semantic invariance, ambiguity-aware correspondence estimation, lack of explicit reasoning during inference, limited cross-domain generalization, poor modeling of dynamic scenes, weak integration with 3D foundation models, and the high cost of global matching. These issues collectively point toward the need for more robust, probabilistic, and scalable matching frameworks.
\end{itemize}

The remainder of this survey is organized as follows. Section \ref{Taxonomy} introduces the proposed taxonomy and provides a comprehensive review of existing methods, structured according to the taxonomy, including {feature extraction}, {single-type feature matcher}, {multi-type feature matcher}, {VFMs based methods}, {training strategy}, and {robust estimation}. Section \ref{Benchmarking} summarizes commonly used datasets, evaluation metrics, and benchmark results, followed by a comparative analysis of representative approaches. Section \ref{Trends} discusses open problems and future research directions. Finally, Section \ref{Conclusion} concludes the survey.

\section{Taxonomy}
\label{Taxonomy}
\begin{figure}[t]
    \centering
    \includegraphics[width=0.48\textwidth]{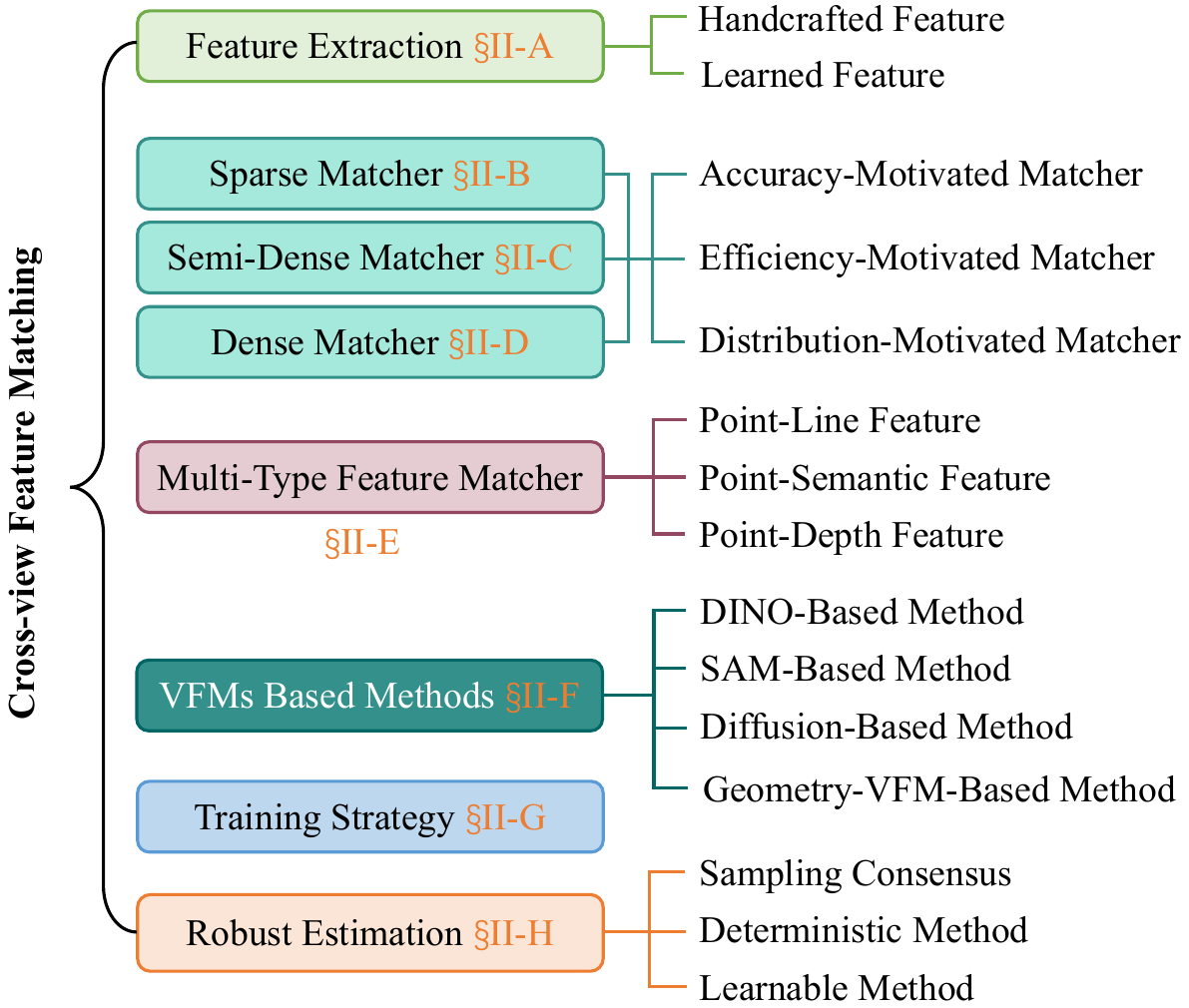}
    \caption{Taxonomy of cross-view feature matching methods.}
    \label{fig:Taxonomy}
\end{figure}

We propose a structured and unified taxonomy of cross-view feature matching methods, aiming to bridge the gap between traditional handcrafted feature matching pipelines and learning-based frameworks. As illustrated in Fig. \ref{fig:Taxonomy}, the proposed taxonomy organizes existing approaches along six key dimensions: \textit{feature extraction}, \textit{single-type feature matcher} (including \textit{sparse matching}, \textit{semi-dense matching} and \textit{dense matching}), \textit{multi-type feature matcher}, \textit{VFMs based methods}, \textit{training strategy}, and \textit{robust estimation}. Within each dimension, we further categorize methods based on their primary design motivations, such as \textit{accuracy}, \textit{efficiency}, \textit{generalization}, or \textit{distribution awareness}, thereby revealing the inherent trade-offs between modeling capacity, computational cost, and applicability across different scenarios. By providing this coherent hierarchical framework, this section not only offers a systematic overview of the methodological landscape, but also lays the groundwork for fair comparisons and in-depth performance analyses.

\subsection{Feature Extraction}
The performance of image matching is fundamentally determined by the quality of feature detection and description, and existing approaches can be broadly categorized into two groups: traditional \textit{handcrafted methods} and \textit{learned methods}. This section provides a comprehensive overview of the evolution of feature detectors and descriptors in image matching, from traditional handcrafted methods to learning-based approaches. Each method has its strengths and is suitable for different applications depending on the specific requirements of the task at hand.

\subsubsection{Feature Detectors}
Feature detectors play a critical role in visual correspondence by identifying stable and repeatable keypoints that serve as anchors for matching across images. Their evolution reflects a transition from handcrafted designs based on geometric and intensity heuristics to learned models that directly optimize repeatability and discriminability in a data driven manner. This progression indicates a broader shift toward unified and trainable frameworks that increasingly balance detection accuracy, robustness to complex variations, and computational efficiency for large scale and real time applications.

\textbf{Handcrafted Feature Detectors.} Handcrafted feature detectors evolved from classical corner-based methods such as Harris \cite{harris1988combined} to scale- and rotation-invariant approaches including SIFT \cite{sift}, SURF \cite{surf}, and MSER \cite{matas2004robust}. Efficiency-oriented detectors such as FAST \cite{FAST} and AGAST \cite{Mair} enabled real-time keypoint extraction through accelerated and adaptive decision trees. Subsequent scale-space methods, including KAZE \cite{KAZE2012} and AKAZE \cite{AKAZE2013}, introduced nonlinear diffusion to improve edge preservation and localization while reducing computational cost through efficient diffusion schemes. BRISK \cite{Leutenegger2011} further combined scale-space detection with efficient binary descriptors, offering a lightweight solution with scale and rotation invariance. Collectively, these methods established the foundations of robust and efficient handcrafted feature detection, balancing invariance, localization accuracy, and computational complexity.

The handcrafted feature detectors have progressively evolved toward achieving a balance between efficiency, invariance, and robustness through carefully designed geometric and intensity-based principles. These methods are typically lightweight and interpretable, making them well-suited for real-time and resource-constrained scenarios. Meanwhile, their limitations in handling complex appearance variations suggest a need for more adaptive and expressive representations.

\textbf{Learned Feature Detectors.} Learned feature detectors have evolved from early end-to-end detection-description frameworks such as LIFT \cite{LIFT2016} and illumination-invariant detectors such as TILDE \cite{TILDE2015} toward unified, data-driven representations. SuperPoint \cite{superpoint} and MagicPoint \cite{superpoint} leverage synthetic pretraining and self-supervision, while D2-Net \cite{D2Net2019}, R2D2 \cite{r2d2}, and KeyNet \cite{KeyNet1,KeyNet2} jointly optimize detection and description, emphasizing dense representation, reliability, and scale-aware repeatability. Lightweight designs such as ALIKE \cite{ALIKE2023} further improve localization accuracy and efficiency, whereas Quad-style networks \cite{QuadNet} exploit unsupervised saliency ranking. DISK \cite{tyszkiewicz2020disk} adopts reinforcement learning to directly optimize correspondence quality, while DeDoDe \cite{edstedt2024dedode} and DaD \cite{edstedt2025dad} increasingly decouple detection from descriptors through SfM-based geometric supervision and descriptor-free self-supervised learning, respectively. The learned detectors have shifted from handcrafted heuristics toward unified and lightweight learning paradigms, with increasing emphasis on geometric consistency, robustness, and computational efficiency for large-scale matching and real-time applications.

\subsubsection{Feature Descriptors}
Feature descriptors constitute a fundamental component of visual correspondence, serving as the basis for establishing reliable matches across images. Their development reflects a continuous evolution from handcrafted representations designed to balance invariance, distinctiveness, and efficiency to learned embeddings that directly optimize discriminability through data driven objectives. This progression highlights a paradigm shift from manual feature engineering to unified learning frameworks that increasingly integrate geometric constraints, contextual information, and task specific optimization for robust and scalable matching.

\textbf{Handcrafted Feature Descriptors.} Handcrafted descriptors have evolved from distinctive, invariant histogram-based representations to compact binary encodings emphasizing computational and memory efficiency. SIFT \cite{sift} and SURF \cite{surf} established robust scale- and rotation-invariant gradient-based descriptors, while PCA-SIFT \cite{PCA-SIFT} and GLOH \cite{GLOH} further improved compactness and distinctiveness through dimensionality reduction and spatial reorganization. Binary descriptors such as BRIEF \cite{brief}, BRISK \cite{Leutenegger2011}, and FREAK \cite{FREAK} significantly reduce computational and storage costs, with BRISK and FREAK additionally providing improved scale and rotation robustness. LIOP \cite{LIOP} instead exploits local intensity-order information to enhance robustness to monotonic illumination changes and image degradation. Subsequent quantized and binarized SIFT/GLOH variants further reduce memory and matching costs through descriptor compression and hashing, trading limited accuracy for substantial efficiency gains \cite{rethinksGLOH}.

The handcrafted feature descriptors have evolved from high-dimensional, gradient-based representations toward compact and efficient encodings that balance distinctiveness, invariance, and computational cost. While earlier designs emphasize robustness to geometric and photometric transformations, later developments focus on lightweight representations and fast matching, making them suitable for real-time and resource-limited applications. However, the handcrafted descriptors are increasingly challenged by complex visual variations and large-scale data scenarios.

\textbf{Learned Feature Descriptors.} Learned descriptors have evolved from CNN-based metric learning toward geometry- and context-aware representations, while increasingly emphasizing compactness and efficiency. Early methods such as DeepDesc \cite{DeepDesc}, L2-Net \cite{L2-Net}, HardNet \cite{HardNet}, and Kumar \emph{et al.} \cite{Kumar2016} established discriminative descriptors through Siamese/triplet architectures and metric-learning objectives. Subsequent approaches incorporate geometric constraints \cite{GeoDesc}, ranking-based objectives \cite{DOAP}, and contextual information \cite{ContextDesc}, while HyNet \cite{HyNet} and SOSNet \cite{SOSNet} further exploit improved similarity modeling and higher-order relational structure. Recent methods increasingly target resource-efficient deployment, with ZippyPoint \cite{kanakis2023zippypoint} introducing compact binary descriptors and XFeat \cite{potje2024xfeat} enabling lightweight sparse and semi-dense matching. The learned descriptors have progressed from patch-level metric learning toward geometry/context-aware, relation-preserving, and increasingly efficient representations.

Feature descriptors have progressed from handcrafted gradient- or binary-based representations to learned embeddings optimized via metric learning and geometric constraints. Recent approaches increasingly incorporate geometric and contextual information to enhance matching performance. Future directions include developing compact yet expressive representations, integrating multi-view geometric consistency, and jointly optimizing descriptors within end-to-end matching frameworks.

\begin{figure*}[t]
    \centering
    \includegraphics[width=\textwidth]{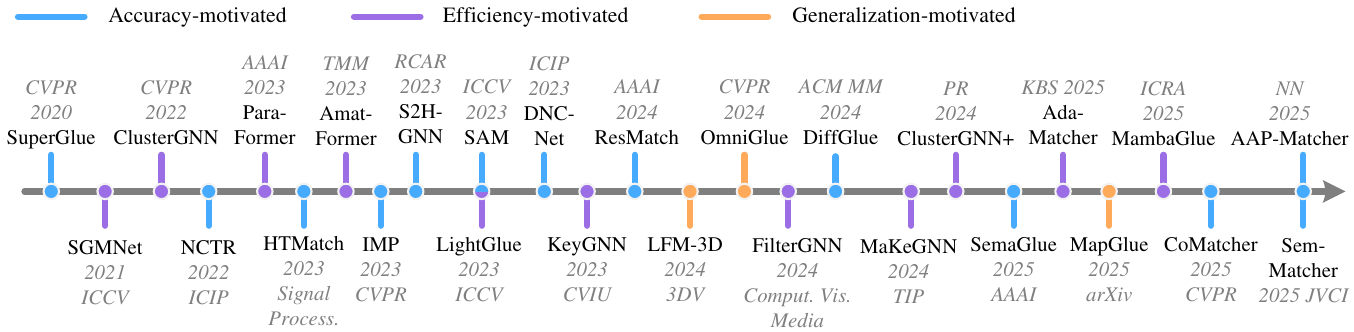}
    \caption{Taxonomy and timeline of sparse matching methods categorized by their motivation (Section \ref{sparse matching sec}).}
    \label{sparse matching}
\end{figure*}

\subsection{Sparse Matcher}
\label{sparse matching sec}
Sparse matching establishes correspondences between two sets of keypoints based on pre-extracted descriptors and spatial information. Before learning-based methods, mutual nearest neighbor (MNN) matching \cite{sift} was the dominant strategy, which relies on pairwise descriptor similarity but lacks inter-image feature interaction. Recent sparse matchers address this limitation by introducing learnable networks to exchange information between feature sets before final correspondence estimation. As shown in Fig. \ref{sparse matching}, existing learning-based sparse matchers can be categorized into three groups according to their primary motivations: \textit{accuracy}, \textit{efficiency}, and \textit{generalization}.

\subsubsection{Accuracy-Motivated Sparse Matcher}
Accuracy remains the primary objective of sparse matching. SuperGlue \cite{superglue} pioneers Transformer-based sparse matching by applying self- and cross-attention to model intra- and inter-image feature interactions, significantly outperforming MNN matching.
Subsequent methods further improve matching accuracy by incorporating additional geometric, contextual, and semantic cues. NCTR \cite{nctr} introduces locality-aware attention through neighbor consensus, while DNC-Net \cite{dncnet} combines sparse neighborhood consensus with dense local consensus from 4D correlation maps. IMP \cite{imp} jointly optimizes correspondence estimation and camera pose prediction, enabling mutual refinement. SAM \cite{sam} exploits scene awareness by predicting keypoint matchability and group-level matching confidence. S2H-GNN \cite{s2hgnn} and HTMatch \cite{htmatch} enhance feature interaction through graph-based aggregation and hybrid Transformer attention, respectively. DiffGlue \cite{diffglue} formulates matching as a diffusion process to iteratively refine the assignment matrix. ResMatch \cite{deng2024resmatch} jointly models descriptor similarity and spatial relationships using enhanced attention mechanisms. SemaGlue \cite{semaglue} and SemMatcher \cite{semmatcher} incorporate semantic information to improve semantic-geometric consistency, while CoMatcher \cite{comatcher} extends sparse matching to multi-view scenarios by exploiting complementary information across views. AAPMatcher \cite{aapmatcher} introduces adaptive pruning and hybrid attention to improve matching quality with reduced redundant interactions.

These accuracy-oriented methods substantially improve correspondence quality through stronger context modeling and richer priors. However, they typically require heavy computation and remain dependent on keypoint detectors, resulting in limited scalability and robustness. Balancing matching accuracy, efficiency, and detector independence remains an important future direction.

\subsubsection{Efficiency-Motivated Sparse Matcher}
Transformer-based sparse matchers often suffer from $\mathcal{O}(N^2)$ complexity due to dense attention over keypoints. Efficiency-oriented methods therefore focus on reducing redundant feature interactions.
SGMNet \cite{chen2021sgmnet} reduces attention complexity by performing graph matching on selected seed points. KeyGNN \cite{keygnn} further improves seed-based aggregation through guided attentional pooling. ClusterGNN \cite{ClusterGNN} reduces computation by partitioning graphs into clusters, while ClusterGNN+ \cite{clustergnn+} extends this idea with coarse-to-fine graph construction and global tokens. ParaFormer \cite{ParaFormer} accelerates matching through parallel self- and cross-attention with shared parameters. LightGlue \cite{lightglue} introduces adaptive early stopping and progressive pruning based on input difficulty. AMatFormer \cite{amatformer} uses anchor features as efficient information bottlenecks, and LightSGM \cite{lightsgm} combines seed selection, graph pooling, and early stopping for lightweight matching. FilterGNN \cite{filtergnn} employs cascaded filtering and linear attention to reduce graph learning complexity. MaKeGNN \cite{makegnn} focuses computation on informative keypoints through matchability-guided sampling. MambaGlue \cite{MambaGlue} adopts Mamba-based sequence modeling for efficient global-local feature aggregation. Ada-Matcher \cite{adamatcher} further improves efficiency through adaptive weight sharing and mask attention.

Efficiency-oriented sparse methods effectively reduce computational cost through pruning, clustering, and adaptive computation. Nevertheless, most approaches still rely on approximations of conventional attention mechanisms, and more fundamental designs for scalable feature interaction remain desirable.

\subsubsection{Generalization-Motivated Sparse Matcher}
Despite strong benchmark performance, sparse matchers often suffer from domain shifts in unseen environments. Recent studies have therefore explored improving generalization across domains and modalities.
OmniGlue \cite{OmniGlue} enhances robustness by constructing keypoint graphs with DINOv2 features and introducing position-guided attention to reduce dependence on spatial cues. LFM-3D \cite{lfm3d} incorporates noisy 3D priors, including monocular depth and normalized object coordinates, to improve correspondence estimation across diverse synthetic and real-world datasets. MapGlue \cite{mapglue} develops a universal multimodal matching framework for remote sensing, using semantic-aware dual-graph guidance to achieve robust cross-modal alignment.

Generalization-oriented sparse matchers improve robustness by leveraging semantic, geometric, and multimodal information. However, compared with accuracy- and efficiency-driven methods, this direction remains relatively underexplored. Future research may focus on learning more domain-agnostic representations with fewer assumptions about specific environments.

\subsection{Semi-Dense Matcher}
\label{semi-dense matching sec}
\begin{figure*}[thb]
    \centering
    \includegraphics[width=\textwidth]{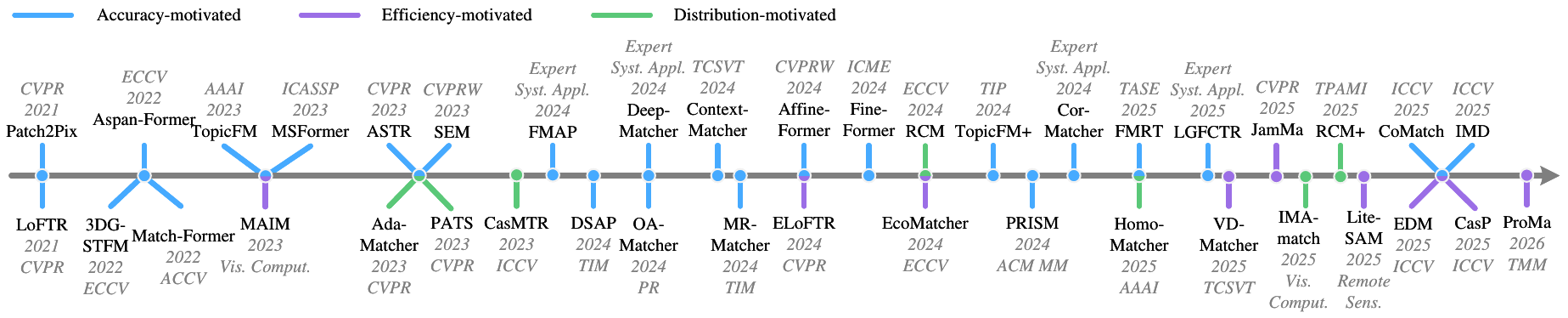}
    \caption{Taxonomy and timeline of semi-dense matching methods categorized by their motivation (Section \ref{semi-dense matching sec}).}
    \label{semi-dense matching}
\end{figure*}

Semi-dense methods follow a coarse-to-fine strategy, first establishing correspondences on low-resolution feature maps and then refining them at higher resolutions. Compared with sparse methods, they provide grid-based coverage and are more effective in texture-less regions. However, dense feature interaction introduces considerable computational costs, and grid-based correspondence generation may limit flexibility in aligning with salient regions. As shown in Fig. \ref{semi-dense matching}, semi-dense matching methods are categorized into three groups according to their primary motivations: \textit{accuracy}, \textit{efficiency}, and \textit{distribution}.

\subsubsection{Accuracy-Motivated Semi-Dense Matcher}
Accuracy-oriented semi-dense methods mainly improve correspondence quality through stronger context modeling, multi-scale feature interaction, and geometric reasoning.
Patch2Pix \cite{patch2pix} introduces a detect-to-refine framework that progressively refines patch-level matches into pixel-level correspondences. LoFTR \cite{loftr} establishes the first semi-dense Transformer matcher, performing coarse matching and fine refinement with dense feature interaction. MatchFormer \cite{matchformer} further develops a fully Transformer-based architecture with interleaved self- and cross-attention. ASpanFormer \cite{aspanformer} introduces adaptive span attention to dynamically adjust receptive fields according to uncertainty.
Subsequent methods enhance matching through additional priors and feature representations. 3DG-STFM \cite{3dgstfm} transfers 3D correspondence knowledge from a multimodal teacher model, while MSFormer \cite{msformer} and TopicFM \cite{topicfm} improve robustness through multi-scale modeling and semantic topic representation, respectively. TopicFM+ \cite{Giangtopicfmplus} further improves efficiency through pooled attention. ASTR \cite{astr}, SEM \cite{sem}, and AffineFormer \cite{affineformer} exploit geometric constraints, depth cues, and transformation modeling to improve consistency. FineFormer \cite{fineformer}, PRISM \cite{prism}, ContextMatcher \cite{contextmatcher}, and DeepMatcher \cite{deepmatcher} enhance multi-scale context aggregation and correspondence refinement through improved attention and matching strategies. FMAP \cite{fmap}, OAMatcher \cite{oamatcher}, and DSAP \cite{dsap} further optimize feature interaction using graph aggregation, overlap-aware matching, and adaptive attention mechanisms. CorMatcher \cite{cormatcher} leverages reliable corner features for feature propagation, while MR-Matcher \cite{mrmatcher} and FMRT \cite{fmrt} improve multi-scale representation and positional modeling. WinMRSI \cite{winmrsi} extends semi-dense matching to cross-modal remote sensing scenarios, LGFCTR \cite{lgfctr} combines CNN and Transformer features, and CoMatch \cite{comatch} improves aggregation efficiency through covisibility-guided token condensation. IMD \cite{imd} formulates matching as a diffusion process to enhance cross-image feature interaction.

Accuracy-motivated semi-dense methods achieve strong performance by exploiting dense context and rich priors. However, their dependence on dense representations and complex architectures leads to high computational costs. Future research may explore more flexible correspondence generation strategies beyond conventional grid-based matching.

\subsubsection{Efficiency-Motivated Semi-Dense Matcher}
Although Transformer-based semi-dense matchers provide strong modeling capability, their computational complexity limits deployment on high-resolution images. Efficiency-oriented methods aim to reduce redundant interactions while preserving matching performance.
EcoMatcher \cite{ecomatcher} introduces Context Clusters for efficient feature aggregation through implicit clustering. ELoFTR \cite{eloftr} reduces computation through adaptive token selection and aggregated attention. MAIM \cite{maim} adopts a Mixer-based architecture to exchange spatial and channel information with low computational overhead. JamMa \cite{jamma} leverages Mamba-based sequence modeling with efficient cross-view scanning and local aggregation. EDM \cite{edm} combines lightweight CNN feature extraction, correlation aggregation, and efficient regression for subpixel matching. CasP \cite{casp} reduces search complexity by establishing coarse correspondence priors before fine matching. VD-Matcher \cite{vdmatcher} improves efficiency through weight reuse and lightweight feature detection. LiteSAM \cite{litesam} develops a lightweight UAV localization matcher by integrating efficient multi-scale extraction, global-local fusion, and subpixel refinement. ProMa \cite{ProMa} systematically rebalances the accuracy-latency trade-off by addressing previously overlooked efficiency bottlenecks through progressive matchability estimation, a hybrid feature interaction module, and feature pruning during coarse matching.

Efficiency-oriented semi-dense methods achieve a better balance between accuracy and computational cost, making them suitable for practical applications. However, simplified architectures may reduce robustness under extreme viewpoint or appearance variations. Improving robustness while maintaining efficiency remains an important direction.

\begin{figure*}[htb]
    \centering
    \includegraphics[width=0.67\linewidth]{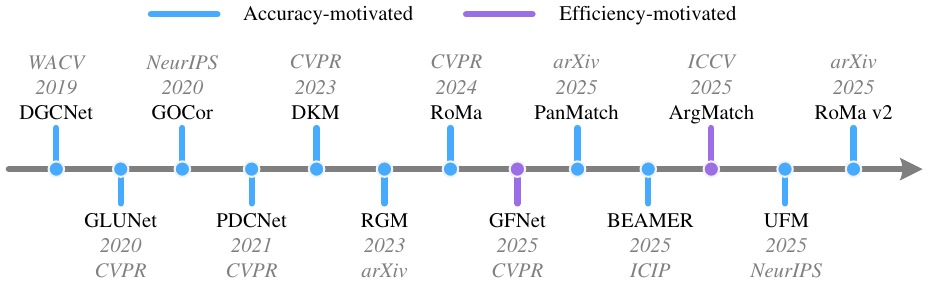}
    \caption{Taxonomy and timeline of dense matching methods categorized by their motivation (Section \ref{dense matching sec}).}
    \label{dense matching}
\end{figure*}

\subsubsection{Distribution-Motivated Semi-Dense Matcher}
Beyond improving matching accuracy, recent studies emphasize the spatial distribution of correspondences, which directly affects downstream tasks such as pose estimation and reconstruction. Conventional semi-dense methods are constrained by one-to-one coarse matching and fixed grids, limiting flexibility under large scale variations. Distribution-oriented methods therefore explore adaptive and flexible correspondence assignment.
PATS \cite{pats} introduces patch-based transportation to model spatially varying scales and enable many-to-many matching. AdaMatcher \cite{adamatcher2} estimates co-visible regions and performs adaptive assignment for many-to-one or one-to-many correspondence. RCM \cite{rcm} improves alignment with salient regions through keypoint-guided matching and scale-aware view switching. CasMTR \cite{casmtr} mitigates coarse-grid limitations using cascade refinement and non-maximum suppression. HomoMatcher \cite{homomatcher} generates fine correspondences by projecting transformed grids rather than relying only on coarse patch centers. IMAmatch \cite{imamatch} selectively increases matching density in important regions through adaptive grid expansion. RCM+ \cite{Lurcm} further introduces a free-form matching paradigm that removes dependence on fixed positional priors and supports arbitrary input structures.

Distribution-motivated semi-dense methods extend conventional matching by improving spatial flexibility and correspondence coverage. However, this direction remains relatively underexplored. Future studies may investigate task-aware matching distributions that optimize correspondence layouts for downstream vision tasks.

\subsection{Dense Matcher}
\label{dense matching sec}
Dense matching methods directly estimate dense correspondence fields between image pairs, often accompanied by confidence prediction to filter unreliable matches. Compared with sparse and semi-dense methods, they provide complete pixel-level correspondence but usually require higher computational and memory costs. As shown in Fig. \ref{dense matching}, existing dense matchers are mainly categorized into two groups according to their motivations: \textit{accuracy} and \textit{efficiency}.

\subsubsection{Accuracy-Motivated Dense Matcher}
Accuracy-oriented dense methods aim to improve correspondence quality through stronger feature representation, geometric modeling, and uncertainty estimation.
Rocco \textit{et al.} \cite{rocco2017convolutional} introduced an end-to-end CNN-based framework that jointly learns feature extraction, correspondence estimation, and geometric transformation fitting. DGC-Net \cite{dgcnet} further incorporated matchability prediction to remove unreliable correspondences, while GLU-Net \cite{glunet} combined global and local correlation mechanisms to handle large displacements and fine motion simultaneously. GOCor \cite{gocor} improved dense correspondence estimation by learning optimization-based correlation volumes with spatial priors. PDC-Net \cite{pdcnet} introduced probabilistic flow estimation with uncertainty prediction for robust matching. DKM \cite{dkm} combined kernel regression, feature refinement, and balanced correspondence sampling for dense matching.

Recent methods further improve robustness through large-scale learning and foundation representations. RGM \cite{rgm} adopts hierarchical learning with uncertainty-based sparsification for robust generalization. RoMa \cite{roma} integrates DINOv2 features, ConvNet-based refinement, and Transformer decoding to achieve robust wide-baseline matching. PanMatch \cite{panmatch} develops a unified correspondence foundation model that leverages large vision model features for zero-shot cross-domain matching. BEAMER \cite{beamer} improves robustness by maintaining multiple correspondence hypotheses through beam-search-based cross-attention. UFM \cite{ufm} jointly learns optical flow and wide-baseline matching using a unified Transformer framework. RoMa v2 \cite{romav2} further improves multi-view correspondence through enhanced warp-correlation learning, efficient refinement, and pixel-wise uncertainty modeling.

Accuracy-motivated dense methods achieve strong performance by exploiting dense representations and large-scale training. However, their high computational and memory requirements restrict practical deployment. Future research may focus on more scalable dense formulations or hybrid strategies that preserve dense matching capability with lower overhead.

\subsubsection{Efficiency-Motivated Dense Matcher}
Dense matching remains computationally demanding due to pixel-level correspondence estimation. Recent efficiency-oriented methods attempt to reduce this burden while maintaining competitive accuracy.
GFNet \cite{gfnet} combines DINOv2 features with a lightweight feature pyramid network and introduces an efficient grid-flow formulation with iterative refinement for accurate and low-cost matching. ArgMatch \cite{argmatch} proposes an adaptive refinement framework that reduces expensive operations through content-aware offset prediction, locally consistent refinement, and adaptive gating.

Compared with accuracy-driven approaches, efficiency-oriented dense matching remains relatively underexplored. Existing studies demonstrate the feasibility of reducing computational costs without significantly sacrificing performance. Future directions may investigate more efficient dense correspondence formulations and hybrid architectures that balance dense modeling capability with practical scalability.

\subsection{Multi-Type Feature Matcher}
\label{muti type}
\begin{figure*}[htb]
    \centering
    \includegraphics[width=0.85\linewidth]{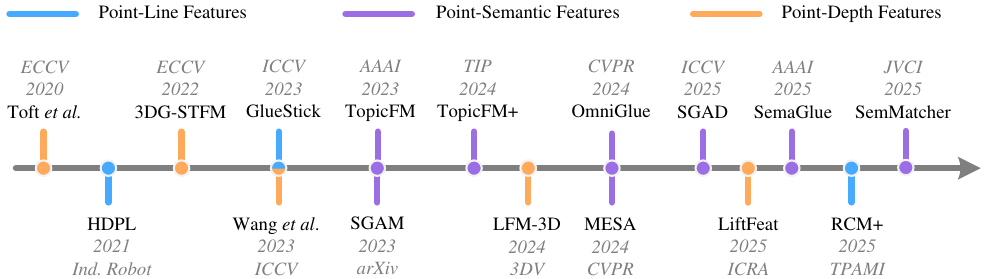}
    \caption{Taxonomy and timeline of multi-type feature matchers (Section \ref{muti type}).}
    \label{muti-type}
\end{figure*}

Traditional geometric matching methods mainly rely on local appearance descriptors, which often degrade under large viewpoint variations, illumination changes, occlusion, blur, and texture deficiency. To improve geometric robustness, recent methods incorporate complementary feature types, such as line segments, semantic cues, and depth information, into unified matching frameworks. As shown in Fig. \ref{muti-type}, existing multi-type feature matchers can be categorized into three groups: \textit{point-line}, \textit{point-semantic}, and \textit{point-depth} features.

\subsubsection{Point-Line Features}
Point-line matching methods extend point-based correspondence by incorporating structural information from line segments. HDPL~\cite{HDPL} represents points and lines as graph nodes and fuses their visual and geometric relationships through graph convolution, improving matching in structured and texture-poor scenes. GlueStick~\cite{gluestick} further unifies keypoints and line segments within a single graph neural network, enabling mutual interaction between heterogeneous geometric primitives through attention and line-specific message passing. RCM+~\cite{Lurcm} introduces a free-form matching paradigm that removes predefined positional constraints and supports arbitrary inputs, including points, edges, and lines, through a position-agnostic encoder and parameter-free decoder.

Point-line methods demonstrate the effectiveness of combining complementary geometric primitives. Future research may extend this idea beyond line segments toward richer structures, such as curves, junctions, and planar elements, for more comprehensive correspondence estimation.

\subsubsection{Point-Semantic Features}
Point-semantic methods integrate high-level semantic information into local feature matching to improve robustness under viewpoint, illumination, and domain variations.
TopicFM \cite{topicfm} introduces a semantic-aware detector-free framework by representing images as latent topic distributions and using semantic regions to constrain feature alignment. TopicFM+ \cite{Giangtopicfmplus} further improves efficiency through pooling and merging attention for semantic-guided fine matching. SGAM~\cite{zhang2023searching} pioneers area-to-point matching by restricting correspondence search within semantically consistent regions, while MESA~\cite{MESA} replaces explicit semantic labels with SAM~\cite{kirillov2023segment}-generated segments and performs robust area matching through multi-relational graph modeling. OmniGlue~\cite{OmniGlue} exploits DINOv2 features to construct semantic-guided keypoint graphs, improving zero-shot generalization across unseen domains. SemaGlue~\cite{semaglue} and SemMatcher \cite{semmatcher} incorporate semantic information into local correspondence through semantic-geometric feature fusion and semantic region constraints, respectively. SGAD~\cite{liu2025sgad} further develops descriptor-based area matching by jointly encoding semantic content and geometric layout, providing a lightweight plug-and-play enhancement for local feature matching.

Point-semantic approaches significantly improve robustness and interpretability by introducing semantic priors from segmentation models and vision foundation models. Future research may investigate tighter integration between semantic reasoning and geometric correspondence estimation for more efficient end-to-end matching.

\subsubsection{Point-Depth Features}
Point-depth methods exploit monocular depth and derived 3D geometric cues to improve matching robustness under viewpoint changes and weak-texture conditions.
Toft \textit{et al.}~\cite{toft2020single} leverage monocular depth to rectify planar regions before feature extraction, reducing viewpoint-induced distortions. 3DG-STFM~\cite{3dgstfm} transfers depth-aware matching knowledge from RGB-D teacher models to RGB-only students through knowledge distillation. LFM-3D~\cite{lfm3d} incorporates normalized object coordinates and monocular depth into graph-based matching to improve wide-baseline correspondence. Wang \textit{et al.}~\cite{wang2023guiding} introduce curvature consistency from depth maps as an additional geometric constraint that can be integrated into existing matching pipelines. LiftFeat~\cite{liftfeat} enhances local descriptors by fusing surface normal features derived from monocular depth through a lightweight geometry-aware lifting module.

Point-depth methods improve matching by introducing 3D geometric priors into 2D correspondence learning. Future research may focus on more accurate monocular geometry estimation and adaptive 2D-3D feature fusion mechanisms to further enhance generalization across diverse scenes.

\begin{figure*}[htb]
    \centering
    \includegraphics[width=\linewidth]{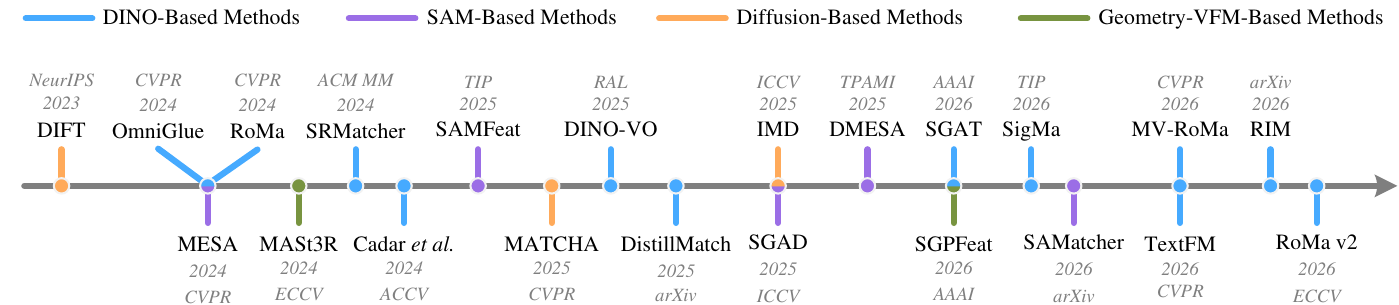}
    \caption{Timeline of VFM based image matching approaches (Section \ref{sec:VFM}).}
    \label{fig;vfm}
\end{figure*}

\subsection{Vision Foundation Models Based Matcher}
\label{sec:VFM}

Recent vision foundation models (VFMs), trained on large-scale datasets, have demonstrated strong
generalization ability and the capability to encode semantic, structural, generative, and geometric
information. These properties provide new opportunities for improving image matching under challenging
conditions, including large viewpoint variations, illumination changes, weak textures, and cross-modal
appearance differences. Instead of replacing dedicated matching networks, existing approaches mainly
exploit VFMs as coarse descriptors, semantic priors, teacher models, or geometry-aware feature
extractors, while retaining specialized modules for accurate correspondence estimation.

As shown in Fig. \ref{fig;vfm}, existing methods can be categorized into four groups:
\textit{DINO-based}, \textit{SAM-based}, \textit{diffusion-based}, and
\textit{geometry-VFM}-based methods, according to the primary VFM used.

\subsubsection{DINO-Based Methods}
The DINO family learns general-purpose visual representations through large-scale self-supervised
training and exhibits strong semantic discrimination and cross-domain generalization. Early studies
mainly explored directly using frozen DINO features, while recent works further investigate semantic-
geometric fusion, adaptation, distillation, and multi-view correspondence construction.

OmniGlue \cite{OmniGlue} combines SuperPoint descriptors with DINOv2 features, using DINOv2-based
semantic cues to guide keypoint information interaction and improve generalization across unseen
domains. RoMa \cite{roma} adopts frozen DINOv2 features as coarse representations and integrates them
with spatially accurate CNN features, establishing a widely used coarse-semantic and fine-geometric
matching paradigm. Based on this idea, SRMatcher \cite{LiuHHFZWLW24} introduces cross-image semantic
fusion with DINOv2 features to enhance correspondence learning, while Cadar et al.
\cite{cadar2024semantic} incorporate DINOv2-derived semantic reasoning into local descriptors to
improve robustness without sacrificing localization accuracy.

Beyond feature enhancement, DINO representations have also been explored for pose estimation and
matching guidance. DINO-VO \cite{DINO-VO} combines DINOv2 features with lightweight CNN-based
geometric features for visual odometry. SigMa \cite{SigMa} utilizes DINOv2-based semantic similarity
to guide salient point selection, coarse matching, and confidence estimation. SGAT \cite{sgat} further
uses DINOv2 to estimate point saliency and matchability, which guides graph-based feature refinement
rather than directly predicting correspondences.

To reduce the computational cost of foundation models, DistillMatch \cite{distillmatch} transfers
modality-invariant knowledge from DINOv2 and DINOv3 into lightweight networks while preserving
local discriminative information. TextFM \cite{textfm} extends DINO-based matching with CLIP-based
language guidance and LoRA adaptation, enabling vision-language assisted correspondence learning.
RIM \cite{rim} exploits DINOv2 tokens for both retrieval and local matching in cross-domain UAV
localization, forming a unified retrieval-and-matching framework. RoMa v2 \cite{romav2} replaces
DINOv2 with DINOv3 and improves matching performance through enhanced dense representations and
architectural refinement. MV-RoMa \cite{mvroma} further extends pairwise RoMa matching to multi-view
track reconstruction by enforcing cross-view consistency.

The DINO-based methods demonstrate strong advantages in semantic representation and cross-domain
generalization. However, their features remain relatively coarse and may confuse visually similar
instances or lose fine-grained spatial details. Therefore, most approaches rely on geometric feature
fusion, task-specific adaptation, or knowledge distillation to bridge the gap between semantic
robustness and precise correspondence localization.

\subsubsection{SAM-Based Methods}

Unlike DINO, which provides dense visual representations, SAM~\cite{kirillov2023segment} mainly offers category-agnostic
regions, accurate boundaries, and hierarchical structural priors. Therefore, SAM-based methods
typically exploit segmentation information to identify coherent or potentially co-visible regions
and use them to facilitate fine-grained correspondence estimation.

MESA \cite{MESA} introduces SAM-based region proposals and organizes them into a multi-relational
area graph, where graph optimization is used to identify reliable matching regions before applying
point-level correspondence methods. DMESA \cite{dmesa} improves the efficiency of MESA by replacing
graph optimization and repeated region matching with dense area-correspondence estimation based on
patch matching and probabilistic modeling. SAMFeat \cite{samfeat} adopts a teacher-student framework
that distills SAM's category-agnostic semantic and boundary knowledge into a lightweight local
feature network, avoiding SAM inference during testing.

Recent studies further explore combining SAM regions with semantic and cross-view reasoning. SGAD
\cite{liu2025sgad} integrates SAM-generated hierarchical regions with DINOv2-based semantic features
and local geometric descriptors, enabling direct region matching while reducing redundant regions.
SAMatcher \cite{samatcher} addresses the inconsistency of independently segmented regions across
views by introducing cross-view interaction to predict co-visible masks and constrain correspondence
estimation.

The SAM-based methods effectively reduce background interference and matching redundancy,
particularly in high-resolution scenarios. However, regions generated from single-view segmentation
may not directly correspond across viewpoints, and mask generation introduces additional computation
costs. Consequently, recent approaches increasingly focus on learned region representations and
explicit cross-view co-visibility modeling.

\subsubsection{Diffusion-Based Methods}
Diffusion models learn rich visual structures and multi-scale representations during image
generation, and their intermediate U-Net features have been shown to contain useful information
for pixel-level correspondence. Unlike discriminative VFMs, diffusion-based methods exploit
generative representations to capture fine spatial details and instance-level structures.

DIFT \cite{DIFT} first demonstrates that intermediate diffusion features can serve as general-purpose
pixel descriptors by extracting activations from a frozen diffusion U-Net. Without task-specific
training, these features support geometric, temporal, and semantic correspondence, although their
performance is sensitive to diffusion timesteps, layers, and feature resolutions. MATCHA \cite{matcha}
further combines multi-level diffusion features with DINOv2-based semantic representations to jointly
model geometric, semantic, and temporal correspondence. IMD \cite{imd} addresses the limitation of
independent feature extraction by introducing cross-image interaction into diffusion-based
representations, enabling more effective discrimination between similar instances.

The diffusion-based methods provide detailed spatial representations and strong structural
priors for correspondence estimation. However, their generative pretraining objective is not
explicitly optimized for geometric matching, resulting in high computational costs and unstable
performance due to feature extraction configurations.

\subsubsection{Geometry-VFM-Based Methods}
Geometry foundation models, including DUSt3R \cite{dust3r} and Depth Anything \cite{yang2024depth}, differ from general-purpose
semantic and generative models because they are trained to predict depth, point maps, camera
parameters, or other geometric quantities. They can therefore support correspondence estimation
by reasoning about the underlying 3D scene rather than relying only on appearance features.

MASt3R \cite{leroy2024grounding} builds on DUSt3R and formulates image matching as 3D point-map prediction,
jointly estimating dense point maps, confidence maps, and local descriptors. Correspondences are
recovered either by comparing points in a shared 3D reference frame or by reciprocal nearest-
neighbor descriptor matching, providing strong robustness to large viewpoint changes.
SGPFeat \cite{sgpfeat} transfers knowledge from geometry-aware foundation models to multimodal image
matching through a dual-teacher distillation framework. Depth Anything V2 supplies high-level
semantic and structural priors, while ALIKE provides keypoint and local geometric information,
and both are distilled into a task-specific multimodal feature extractor through heterogeneous
feature aggregation.

The geometry-VFM-based methods provide a direct connection between image correspondence and 3D scene
understanding and are particularly advantageous under substantial viewpoint changes. However, they
are generally more computationally demanding and require large-scale 3D training data. Their
predictions can also become unreliable for domains that differ substantially from the training
data.

\subsection{Training Strategy}
\label{training framework sec}
\begin{figure}[h]
    \centering
    \includegraphics[width=0.85\linewidth]{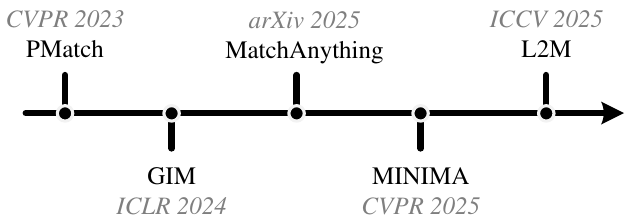}
    \caption{Timeline of typical training strategies (Section \ref{training framework sec}).}
    \label{training framework}
\end{figure}

Beyond improving the matcher itself, some studies focus on developing enhanced training frameworks that utilize larger and more diverse datasets to train more robust and generalizable matchers. As the timeline shown in Fig. \ref{training framework}, PMatch \cite{pmatch} introduces a paired masked image modeling pretext task to jointly pretrain the encoder and decoder, and proposes a cross-frame global matching module that enhances robustness to textureless regions, further strengthened by a homography-based loss tailored for planar structures.
GIM \cite{gim} proposes a self-training framework that leverages internet videos to train a single generalizable image matcher, generating and propagating pseudo-labels without relying on 3D reconstruction for improved efficiency and robustness.
MINIMA \cite{minima} proposes generating large-scale pseudo multimodal datasets from easily accessible RGB images using generative models, enabling unified cross-modal matching across modalities such as infrared, depth, event, normal, and artistic styles while maintaining low cost, high flexibility, and high-quality correspondence labels.
MatchAnything \cite{matchanything} proposes a large-scale universal cross-modality pre-training framework that integrates multi-view geometry, video continuity, and synthetic warping into a unified dataset engine to generate diverse image pairs with reliable ground-truth correspondences for robust and generalizable image matching.
L2M \cite{l2m} proposes a two-stage framework that lifts 2D images into 3D space by first learning a 3D-aware feature encoder through multi-view synthesis and Gaussian feature representation, and then employing novel-view rendering with large-scale synthetic data to train a robust, generalizable feature matcher.

The training frameworks discussed above play a fundamental role in data-driven matching, as improvements in data generation, supervision, and pretraining can benefit a wide range of methods. Existing approaches enhance robustness and generalization by leveraging large-scale data, self-supervision, and multimodal synthesis. Future directions may explore feedback-driven data generation and more resource-efficient training paradigms.

\subsection{Robust Estimation}
\label{robust_estimation_sec}
\begin{figure*}[thb]
    \centering
    \includegraphics[width=\linewidth]{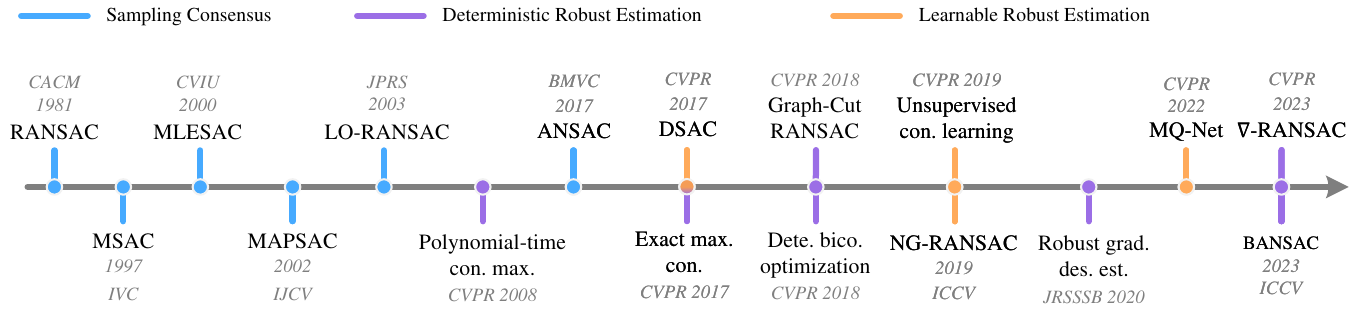}
    \caption{Taxonomy and timeline of robust estimation methods (Section \ref{robust_estimation_sec}).}
    \label{robust_estimation_fig}
\end{figure*}

Robust estimation refers to the problem of estimating model parameters from data that may contain noise, outliers, or deviations from idealized assumptions, such that the estimator remains accurate and stable even when a significant portion of the observations are corrupted \cite{huber1981robust}.
In the context of visual correspondence, robust estimation refers to techniques that reliably estimate geometric models, \textit{e.g.}, homographies, essential/fundamental matrices, or relative camera poses, from putative feature correspondences that are contaminated by noise and a high proportion of outliers \cite{diakonikolas2023algorithmic}. Since feature detectors and descriptors typically yield candidate correspondences with significant mismatches, robust estimation is essential to reject outliers before model fitting. As shown in Fig. \ref{robust_estimation_fig}, we classify existing robust estimation approaches into three categories: \textit{sampling consensus}, \textit{deterministic robust estimation}, and \textit{learnable robust estimation}. These methods improve the accuracy and stability of downstream tasks such as panoramic stitching, visual odometry, and 3D reconstruction.

\subsubsection{Sampling Consensus for Robust Estimation}
Sampling consensus algorithms combine random sampling with the evaluation of consensus sets to identify inlier subsets. The canonical approach in image matching is Random Sample Consensus (RANSAC) \cite{fischler1981random}, which repeatedly samples minimal subsets of point correspondences to hypothesize a model, \textit{e.g.}, homography or fundamental matrix, and then counts inliers that satisfy a geometric error threshold. The model with the largest inlier set is selected as the robust estimate. RANSAC’s ability to tolerate a large fraction of outliers made it a widely used method for robust homography and pose estimation in feature-based matching pipelines.

Variants of RANSAC address its limitations in convergence speed and noise sensitivity \cite{martinez2022ransac}. For instance, MSAC \cite{torr1997robust} is an early RANSAC variant that incorporates maximum likelihood estimation for model parameters, initially applied to trifocal tensor estimation, improving accuracy over standard RANSAC at the cost of higher computational load. MLESAC \cite{torr2000mlesac} generalizes RANSAC by selecting solutions based on maximum likelihood rather than inlier count, and combines it with an automatic parameterization method that handles nonlinear constraints for multiview relations. MAPSAC \cite{torr2002bayesian} extends the MSAC and MLESAC framework by adopting a Bayesian approach to compute a robust Maximum a Posterior (MAP) estimate, enabling least-squares fitting of arbitrary manifolds, including lines and planes, with improved robustness over previous maximum likelihood formulations. Locally Optimized RANSAC (LO-RANSAC) \cite{chum2003locally} improves RANSAC by incorporating a local optimization step, addressing the empirical observation that more samples are often needed than theory predicts, and relaxing the assumption that a model computed from an all-inlier sample must be consistent with the entire inlier set. Adaptive Non-minimal Sample and Consensus (ANSAC) \cite{fragoso2017ansac} uses ranked subsets with estimated inlier ratios to generate higher-quality hypotheses and improve convergence performance in matching tasks. Optimized probabilistic sampling techniques such as BANSAC \cite{BANSAC} that employs dynamic Bayesian networks to perform weighted, adaptive sampling of data points, updating inlier probabilities iteratively and introducing a probabilistic stopping criterion to improve efficiency and robustness.

The sampling consensus methods provide a robust framework for model estimation in the presence of outliers by iteratively sampling minimal subsets and selecting hypotheses with maximum consensus. Subsequent improvements enhance estimation accuracy, convergence speed, and statistical robustness through likelihood modeling, Bayesian inference, and local optimization. Despite these advances, an important open problem remains how to achieve both high computational efficiency and strong robustness under extreme outlier ratios and complex transformations, particularly in real-time matching scenarios.

\subsubsection{Deterministic Robust Estimation}
Recently, efficient deterministic algorithms for consensus maximization have attracted increasing attention. Unlike random sampling methods, deterministic optimization approaches aim to find globally consistent inlier sets or model parameters without reliance on random sampling. These methods often formulate geometric model fitting as an explicit optimization problem that maximizes an inlier count or minimizes a robust objective, and solve it using deterministic search or convex relaxation techniques. Although they do not aim for global optimality, their directed search strategy often yields high-quality solutions. For example, Olsson \textit{et al.} \cite{olsson2008polynomial} proposed a polynomial-time algorithm for computing globally optimal point-set alignments in the presence of outliers by maximizing consensus, using convex programming techniques grounded in computational geometry. Le \textit{et al.} \cite{le2017exact} proposed a deterministic and locally convergent algorithm for maximum consensus estimation based on a penalized linear complementarity formulation and a Frank-Wolfe optimization scheme. Cai \textit{et al.} \cite{cai2018deterministic} introduced an efficient deterministic algorithm that improves an initial solution for consensus maximization by formulating each update as a biconvex optimization problem, enabling reliable and significant consensus improvement without random sampling. Graph-Cut RANSAC \cite{barath2018graph,Graph-Cut} formulates the local optimization step as an energy minimization problem solved via graph cuts, combining unary residuals and a binary term to enforce spatially coherent inlier labeling, and demonstrates superior accuracy, reliability, and efficiency across multiple tasks, such as homography, 6D pose, and fundamental/essential matrix estimation. Prasad \textit{et al.} \cite{prasad2020robust} introduced a computationally efficient class of robust estimators based on a novel gradient descent variant for general convex risk minimization, providing provable robustness under classical $\epsilon$-contamination and heavy-tailed settings, with theoretical guarantees and practical effectiveness demonstrated for linear regression, logistic regression, and exponential family parameter estimation. While computationally more demanding than sampling-based methods, deterministic solvers yield repeatable results with theoretical bounds that can be advantageous in safety-critical vision systems.



\subsubsection{Learnable Robust Estimation}
Learnable robust estimation incorporates machine learning, particularly deep learning, into traditional robust estimation pipelines to enhance outlier rejection and consensus evaluation. These methods train models to predict inlier probabilities or to generate features that facilitate robust matching under challenging conditions such as illumination variation and extreme viewpoint changes. DSAC \cite{2017dsac} introduces a differentiable variant of RANSAC by replacing the traditional hard selection of hypotheses with probabilistic selection, allowing gradients to propagate through the model scoring process. MQ-Net \cite{barath2022learning} differs from prior RANSAC variants by replacing traditional consensus- or residual-based scoring with a learned Model Quality Network for direct error prediction, and by introducing a Minimal Samples Filtering Network to preemptively discard degenerate or geometrically inconsistent hypotheses, thereby improving both efficiency and accuracy. Progressive consensus learning methods \cite{zhao2021progressive} iteratively prune correspondence sets using learned consensus scores derived from local and global graph contexts, improving the separability of inliers and outliers before geometric estimation. Such schemes have demonstrated superior performance over classical consensus mechanisms, particularly in scenarios with strong descriptor ambiguity or high mismatch rates. Neural Guided RANSAC \cite{brachmann2019neural} learns a categorical sampling distribution via an approximate expected gradient, while $\nabla$-RANSAC \cite{wei2023generalized} introduces a fully differentiable RANSAC formulation that employs the Gumbel reparameterization trick to learn the sampling distribution in an end-to-end manner. Probst \textit{et al.} \cite{probst2019unsupervised} introduced the first unsupervised learning framework for consensus maximization by deriving a transformation-free geometric constraint that enables learning to identify the largest inlier set, achieving robust model fitting and outperforming RANSAC on multiple 3D vision tasks.

Learnable robust estimation extends classical consensus-based frameworks by integrating deep learning to guide hypothesis generation, sampling, and inlier evaluation in a data-driven manner. It enables end-to-end optimization by making the sampling and scoring processes differentiable. Despite these advances, a key open problem lies in achieving strong generalization across diverse scenes and geometric configurations, particularly under distribution shifts, while maintaining the interpretability and stability traditionally offered by model-based robust estimation methods.

\section{Benchmarking and Analysis}
\label{Benchmarking}
A unified benchmarking of representative cross-view feature matching methods under consistent evaluation protocols is presented in this section. Through systematic comparisons across sparse, semi-dense, and dense paradigms on various tasks, including relative pose estimation, homography estimation, and visual localization, we reveal the performance trade-offs: dense methods achieve the highest accuracy at greater computational cost, while sparse methods remain efficient but are limited by detector quality. This benchmarking provides fair, quantitative insights into the strengths and limitations of existing approaches.

\begin{figure*}[t]
    \centering
    \includegraphics[width=\textwidth]{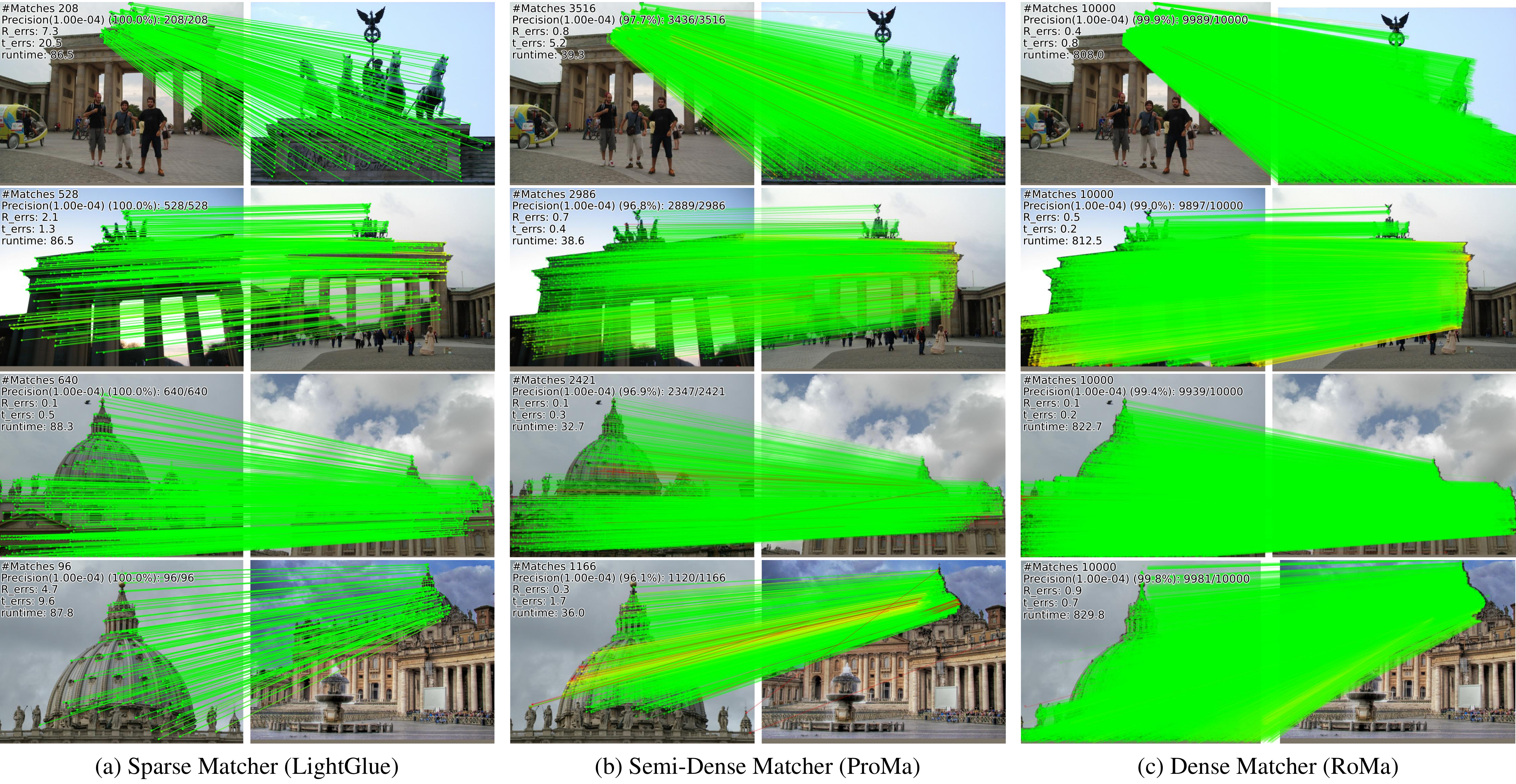}
    \caption{Visualization of typical cross-view image matching results from sparse, semi-dense, and dense matchers.}
    \label{fig:visualization}
\end{figure*}

\subsection{Datasets}

\textbf{YFCC100M} \cite{thomee2016yfcc100m} contains approximately 100 million creative commons-licensed outdoor images from Flickr, along with metadata such as camera parameters, user tags, and partial geolocation. Due to its scale and real-world diversity, it has become a widely adopted benchmark for large-scale computer vision tasks. For image matching, OANet \cite{oanet} selects 72 landmark-centric sequences reconstructed using COLMAP \cite{schonberger2016structure}, which provides sparse 3D models and camera poses as ground-truth references. Typically, 68 sequences are used for training/validation and 4 held out for testing.

\textbf{MegaDepth} \cite{li2018megadepth} provides SfM and Multi-View Stereo (MVS) reconstructions of 196 globally distributed landmark scenes. Using COLMAP \cite{schonberger2016structure} and MVS pipelines, it delivers RGB images, camera parameters, sparse 3D point clouds, and dense depth maps. MegaDepth is widely adopted as a benchmark for wide-baseline image matching and relative pose estimation due to its challenging real-world conditions, including extreme viewpoint variations, illumination changes, and repetitive structures. A standard evaluation protocol, often referred to as MegaDepth-1500 \cite{loftr}, samples 1,500 image pairs from two iconic scenes \textit{Sacre Coeur} and \textit{St. Peter’s Square}.

\textbf{ScanNet} \cite{dai2017scannet} is a large-scale indoor RGB-D dataset containing 1,513 scanned scenes with approximately 2.5 million views. Each view provides an RGB image, a depth map, and ground-truth camera poses obtained via dense 3D reconstruction. The dataset features challenging indoor conditions such as texture scarcity, repetitive structures, and occlusions, making it a standard benchmark for indoor image matching and pose estimation. A common evaluation protocol samples 1,500 image pairs from the test split, as used in multiple matching methods~\cite{superglue}\cite{loftr}.

\textbf{SUN3D} \cite{xiao2013sun3d} is a large-scale indoor RGB-D dataset comprising 254 sequences captured across diverse indoor environments. It provides RGB images, depth maps, and ground-truth relative camera poses refined through generalized bundle adjustment~\cite{hartley2003multiple}. The scenes exhibit challenging conditions for image matching, including sparse textures, repetitive patterns, and self-occlusions. Following common practice~\cite{oanet}, 239 sequences are used for training and validation, while the remaining 15 sequences form the test set.

\textbf{HPatches} \cite{balntas2017hpatches} offers 116 real-world image sequences for evaluating local feature matchers. In each sequence, a reference image is paired with warped views and pixel-accurate homographies between them are provided as ground-truth. The collection is intentionally split into two complementary challenges: 57 sequences capture significant viewpoint changes, while 59 emphasize illumination differences. This separation allows researchers to dissect the geometric versus photometric robustness of both classical and deep learning-based matching approaches.

\textbf{Aachen Day-Night } \cite{sattler2018benchmarking} comprises two versions. The v1.0 version covers the historic center of Aachen which consists 4,328 daytime images and 98 nighttime images. The v1.1 version expands the dataset to 6,697 daytime reference images and 1,015 query images (824 daytime and 191 nighttime). In both versions, ground-truth camera poses are available for query images, enabling quantitative evaluation under challenging conditions such as extreme lighting variations, viewpoint differences, and seasonal effects.

\textbf{InLoc}~\cite{taira2018inloc} is a challenging indoor visual localization benchmark. The dataset comprises 9,972 high-resolution RGB-D reference images, which are precisely aligned with accurate 3D reconstructions and corresponding floorplans. In addition, it provides 329 handheld query images captured using an iPhone 7 under varying lighting and viewing conditions, each annotated with ground-truth 6-DoF camera poses verified through rigorous bundle adjustment. The dataset captures real-world complexities commonly encountered in indoor environments.

\subsection{Evaluation Metrics}

\textbf{Relative Pose Estimation.}
Relative pose estimation is commonly evaluated by the angular errors of translation and rotation between the estimated and ground-truth poses. The pose error is typically defined as the maximum of the two errors, and performance is summarized by the normalized Area Under the Curve (AUC) of the recall curve. Following common practice, we report $\mathrm{AUC}@5^{\circ}$, $\mathrm{AUC}@10^{\circ}$, and $\mathrm{AUC}@20^{\circ}$.

\textbf{Homography Estimation.}
Homography estimation is typically evaluated using the average corner reprojection error between the estimated and ground-truth homographies. The proportion of samples whose reprojection error falls below a given pixel threshold is used to construct the accuracy curve, from which the AUC is computed. We report $\mathrm{AUC}@3$ px, $\mathrm{AUC}@5$ px, and $\mathrm{AUC}@10$ px.

\textbf{Visual Localization.}
Visual localization is evaluated by the translational and rotational errors between the predicted and ground-truth camera poses. A query is considered successfully localized when both errors fall below predefined distance and angular thresholds. The localization accuracy is then measured by the proportion of successfully localized queries under each threshold pair.

\begin{table*}[htb]
	\centering
	\caption{Results of relative pose estimation on MegaDepth Dataset (Outdoor) and ScanNet (Indoor) Dataset.  Sparse methods and the majority of semi-dense methods are evaluated using a single model trained on the MegaDepth dataset. For dense methods, Models are trained separately on each respective dataset. For several semi-dense methods, the source code is not publicly available, making it infeasible to determine their parameter counts. Moreover, their published ScanNet results were obtained using models trained on ScanNet. We denote these results with an underline to indicate this distinction. Additionally, CasMTR and IMAmatch are representative cascaded matching approaches that bridge semi-dense and dense paradigms. They are also evaluated on ScanNet using models trained specifically for that dataset, enabling a direct comparison with dense methods. AUC@$\theta^\circ$ denotes the area under the curve at $\theta^\circ$ threshold.}
	\label{pose}
		\begin{tabular}{llccccccc}
			\toprule
			\multirow{2}{*}{Category} & \multirow{2}{*}{Method} & \multirow{2}{*}{Params (M)} & \multicolumn{3}{c}{MegaDepth Dataset} & \multicolumn{3}{c}{ScanNet Dataset} \\
           \cmidrule(l){4-6} \cmidrule(l){7-9}
			& & & AUC@5$^\circ$ & AUC@10$^\circ$ & AUC@20$^\circ$ & AUC@5$^\circ$ & AUC@10$^\circ$ & AUC@20$^\circ$ \\
			\midrule
			\multirow{8}{*}{Sparse}
			& MNN & 0 & 31.7 & 46.8 & 60.1 & 7.5 & 18.6 & 32.1 \\
            & ParaFormer & 12.5 & 44.5 & 62.0 & 76.2 & 14.4 & 30.4 & 48.0 \\
            & SAM & 13.5 & 45.8 & 63.2 & 76.7 & 14.8 & 31.0 & 48.3 \\
            & SuperGlue & 12.0 & 48.7 & 65.8 & 78.9 & 17.4 & 33.4 & 49.0 \\
            & LightGlue & 11.9 & 49.9 & 67.0 & 80.1 & 18.5 & 35.5 & 51.8 \\
            & MambaGlue & 16.5 & 50.1 & 67.5 & 80.3 & 18.2 & 35.1 & 51.4 \\
            & SemaGlue & 29.2 & 60.5 & 74.2 & 84.1 & 18.9 & 36.2 & 52.2 \\
            & DiffGlue & 21.5 & 49.6 & 67.2 & 80.5 & 17.5 & 34.6 & 51.6 \\
			
			\midrule
			\multirow{18}{*}{Semi-Dense}
			& DRC-Net & 29.2 & 27.0 & 42.9 & 58.3 & 7.7 & 17.9 & 30.5 \\
			& LoFTR & 11.6 & 52.8 & 69.2 & 81.2 & 16.9 & 33.6 & 50.6 \\
            & RCM & 9.8 & 53.2  & 69.4  & 81.5 & 17.3  & 34.6  & 52.1 \\
			& QuadTree & 12.7 & 54.6 & 70.5 & 82.2 & 19.0 & 37.3 & 53.5 \\
			& MatchFormer & 20.3 & 53.3 & 69.7 & 81.8 & 15.8 & 32.0 & 48.0 \\
			& TopicFM & 10.5 & 54.1 & 70.1 & 81.6 & 17.3 & 35.5 & 50.9 \\
			& AspanFormer & 15.8 & 55.3 & 71.5 & 83.1 & 19.6 & 37.7 & 54.4 \\
			& SEM & - & 58.0 & 72.9 & 83.7 & 18.7 & 36.6 & 52.9 \\
			& EcoMatcher & - & 56.5 & 72.0 & 83.4 & \underline{25.8} & \underline{46.2} & \underline{64.0} \\
			& ELoFTR & 16.0 & 56.4 & 72.2 & 83.5 & 19.2 & 37.0 & 53.6 \\
            & JamMa & 5.7 & 56.0 & 71.3 & 82.2 & 11.5 & 25.3 & 40.5 \\
			& TopicFM+  & 11.6 & 58.2 & 72.8 & 83.2 & 20.4  & 38.5 &  54.5 \\
			& EDM  & 10.3 & 57.5 & 73.2 & 84.2 & 19.8  & 37.5 &  54.4 \\
			& CoMatch  & 12.0 & 58.0 & 73.2 & 84.2 & 21.7  & 40.2 &  56.7 \\
			& AffineFormer  & 12.8 & 57.3 & 72.8 & 84.0 & 22.0  & 40.9 &  58.0 \\
			& CasP & 16.3 & 57.1 & 72.7 & 83.9 & 23.0  & 41.6 &  58.7 \\
            & CasMTR & 14.3 & 59.2 & 74.3 & 84.8 & \underline{27.0} & \underline{47.1} & \underline{64.4} \\
			& IMAmatch & 84.1 & 62.8 & 76.6 & 86.2 & \underline{31.5} & \underline{52.7} & \underline{70.2} \\
			\midrule
			\multirow{6}{*}{Dense}
			& PDC-Net+ & 18.4 & 51.5 & 67.2 & 78.5 & 20.3 & 39.4 & 57.1 \\
			& PMatch & 42.5 & 61.4 & 75.7 & 85.7 & 29.4 & 50.1 & 67.4 \\
			& DKM & 72.3 & 60.4 & 74.9 & 85.1 & 29.4 & 50.7 & 68.3 \\
			& RoMa & 111.3 & 62.6 & 76.7 & 86.3 & 31.8 & 53.4 & 70.9 \\
			& UFM & 428.3 & 41.5 & 57.9 & 72.4 & 31.3 & 54.1 & 72.0 \\
			& RoMa v2 & 425.4 & 62.8 & 77.0 & 86.6 & 33.6 & 56.2 & 73.8 \\
			\bottomrule
		\end{tabular}
\end{table*}

\subsection{Relative Pose Estimation}
\textbf{Settings.}
We evaluate representative sparse, semi-dense, and dense methods on MegaDepth \cite{li2018megadepth} and ScanNet \cite{dai2017scannet} for two-view camera pose estimation. We report AUC at $5^\circ$, $10^\circ$, and $20^\circ$. For sparse methods, SuperPoint is uniformly adopted as the keypoint detector. On MegaDepth, input images are resized such that the longer side is 1,600 pixels, with up to 2,048 keypoints retained. On ScanNet, images are kept at their original resolution of
$480 \times 640$ pixels, with up to 1,024 keypoints. For semi-dense methods, which do not rely on explicit keypoint detector, input images are resized to 1,152$\times$1,152 pixels on MegaDepth and kept at $480 \times 640$ pixels on ScanNet. For dense methods, the settings are retained as originally proposed in the respective papers \cite{pdcnet,pmatch,dkm,roma,ufm,romav2}. All methods estimate the relative camera pose using RANSAC, with an error threshold of 0.5 pixels for MegaDepth and 1.0 pixel for ScanNet.

\textbf{Results.}
As reported in Table~\ref{pose}, sparse matching methods represented by SuperGlue and LightGlue achieve strong performance, with AUC@20$^\circ$ mostly exceeding 80\%. This demonstrates that keypoint-based sparse matching remains highly effective in outdoor scenes. Notably, LightGlue achieves an optimal balance between performance and model size, establishing a highly efficient baseline. Recent variants such as SemaGlue, DiffGlue, and MambaGlue further enhance feature representation, achieving improved accuracy. However, it is worth noting that these subsequent models generally adopt a strategy of trading increased parameter count for higher precision. On the weakly textured ScanNet dataset, these methods drop sharply, with AUC@20$^\circ$ typically below 50\%. Fundamentally, this limitation stems from their reliance on the keypoint detector. The upper bound of matching accuracy is constrained by the number, spatial distribution, and descriptor quality of the detected keypoints. Although sparse methods are generally computationally efficient and suitable for real-time applications, this inherent bottleneck cannot be fully compensated by improvements in the matching module alone.

In comparison, semi-dense methods exhibit stronger performance on both datasets. LoFTR was the first to introduce Transformers into semi-dense matching and proposed a coarse-to-fine paradigm. Thanks to the inherent advantage of denser feature correspondences, LoFTR, despite being proposed several years ago, achieves comparable performance to SOTA sparse methods. Subsequent works have primarily improved LoFTR along two directions: enhancing inference efficiency and strengthening feature representation, for example by designing more efficient attention mechanisms or incorporating semantic priors. Moreover, methods such as CasMTR and IMAmatch adopt a cascaded matching strategy, extending the LoFTR framework toward fully dense matching and achieving significant performance gains. Notably, IMAmatch achieves AUC scores of 62.8 / 76.6 / 86.2 on MegaDepth and 31.5 / 52.7 / 70.2 on ScanNet at thresholds of $5^\circ$, $10^\circ$, and $20^\circ$, respectively, reaching a level comparable to that of dense methods. However, this performance parity comes at the cost of model scale. IMAmatch's parameter count approaches that of fully dense matching approaches, reflecting a trade-off between accuracy and model compactness.

The performance of two-view pose estimation methods systematically improves with increasing matching density: dense methods consistently outperform semi-dense approaches, which in turn generally surpass sparse methods. State-of-the-art dense methods, such as RoMa v2, achieve the highest accuracy on both benchmarks. On ScanNet, RoMa v2 attains an AUC@20$^\circ$ of 73.8\%, significantly exceeding all semi-dense and sparse methods. This result clearly demonstrates the advantage of end-to-end dense modeling in handling complex geometry and low-texture environments. However, these methods are often computationally expensive accompanied by a substantial increase in model parameters.

\subsection{Homography Estimation}
\textbf{Settings.}
Homography estimation experiments are conducted on the HPatches \cite{balntas2017hpatches}, following the evaluation protocols from LightGlue \cite{lightglue} and ELoFTR \cite{eloftr}.
For methods that require keypoints detector, images are resized such that the shorter side is 480 pixels, and up to 1,024 keypoints are extracted per image. Semi-dense methods, which do not rely on explicit keypoint detector, are applied on images processed with the same resizing strategy. Dense methods are evaluated using the settings specified in their respective original papers \cite{pdcnet,dkm,pmatch,roma}. Additionally, to ensure fair comparison and follow LoFTR \cite{loftr}, only the top 1,000 predicted matches from semi-dense and dense methods are retained for homography estimation. All methods use RANSAC to estimate the homography, with a threshold of 0.5 for sparse methods and 3.0 for semi-dense and dense methods.

\textbf{Results.}
As reported in Table \ref{tab:homography}, three paradigms exhibit a clear hierarchical pattern in overall performance. Sparse methods are relatively limited in global alignment capability due to constraints on keypoint quantity and distribution. Semi-dense methods expand matching density, achieving noticeable gains in robustness and accuracy. Dense methods model pixel-level correspondences directly and attain the highest overall performance. This hierarchy remains consistent across different evaluation strictness levels. The performance gap among the three paradigms becomes more pronounced under stricter threshold. Overall, finer matching granularity correlates positively with better homography estimation performance.

\begin{table}[tb]
\centering
\caption{Results of homography estimation on HPatches Dataset. All methods are categorized into three groups: sparse, semi-dense, and dense. The AUC of reprojection error of corner points at different thresholds is reported.}
\label{tab:homography}
{\small
\begin{tabular}{clccc}
    \toprule
    \multirow{2}{*}{Category} & \multirow{2}{*}{Method} & \multicolumn{3}{c}{Homography est. AUC} \\
    \cmidrule(lr){3-5}
    & & @3px & @5px & @10px \\
    \midrule
    \multirow{10}{*}{Sparse} & D2Net + NN & 23.2 & 35.9 & 53.6 \\
    & R2D2 + NN & 50.6 & 63.9 & 76.8 \\
    & DISK + NN & 52.3 & 64.9 & 78.9 \\
    & SuperGlue & 56.8 & 68.7 & 80.9 \\
    & ParaFormer & 53.6 & 67.6 & 81.0 \\
    & SAM & 55.2 & 68.9 & 81.5 \\
    & LightGlue & 56.2 & 68.3 & 81.0 \\
    & DiffGlue & 56.2 & 68.3 & 81.0 \\
    & MambaGlue & 58.5 & 70.2 & 81.8 \\
    & SemaGlue & 56.2 & 68.3 & 81.0 \\

    \cmidrule(lr){1-5}
    \multirow{15}{*}{Semi-Dense}
    & DRC-Net & 50.6 & 56.2 & 68.3 \\
    & Patch2Pix & 59.3 & 70.6 & 81.2 \\
    & LoFTR & 65.9 & 75.6 & 84.6 \\
    & RCM & 63.0 & 74.6 & 84.9 \\
    & TopicFM & 67.3 & 77.0 & 85.7 \\
    & ASpanFormer & 67.4 & 76.9 & 85.6 \\
    & Efficient-LoFTR & 66.5 & 76.4 & 85.5 \\
    & JamMa & 68.1 & 77.0 & 85.4 \\
    & EDM & 68.5 & 78.1 & 86.6 \\
    & CoMatch & 68.4 & 78.2 & 86.8 \\
    & EcoMatcher & 68.0 & 77.8 & 86.4 \\
    & CasMTR & 71.4 & 80.2 & 87.9 \\
    & SEM & 69.6 & 79.3 & 87.0 \\
    & ASTR & 71.7 & 80.3 & 88.0 \\
    & IMAmatch & 72.4 & 81.1 & 88.9 \\
    \cmidrule(lr){1-5}
    \multirow{4}{*}{Dense} & PDC-Net+ & 67.8 & 77.5 & 86.2 \\
    & DKM & 71.4 & 80.5 & 88.4 \\
    & PMatch & 71.9 & 80.7 & 88.5 \\
    & RoMa & 72.2 & 81.2 & 89.1 \\
    \bottomrule
\end{tabular}
}
\end{table}

\subsection{Visual Localization}
\textbf{Settings.}
All experiments follow the standard visual localization pipeline implemented in the open-source framework HLoc \cite{sarlin2019coarse}. A sparse 3D reconstruction is employed as the map representation. For Aachen Day-Night \cite{sattler2018benchmarking}, it is built from reference images using COLMAP and their camera intrinsics and poses. For InLoc \cite{taira2018inloc}, the official precomputed point cloud is adopted without modification.
Given a query image, candidate reference views are retrieved using NetVLAD \cite{netvlad}. Feature matching is then performed between the query and each retrieved reference. The resulting correspondences serve as input to a RANSAC-based PnP solver to estimate the camera pose. The inlier threshold in RANSAC is set to 12 pixels for Aachen Day-Night and 48 pixels for InLoc.
Sparse matchers process images resized to 1,024 pixels on the longer side for Aachen Day-Night v1.0 and 1,600 pixels for InLoc, with 4,096 keypoints extracted by SuperPoint. Semi-dense matchers operate at 1,024 pixels on the long edge for Aachen Day-Night v1.1 and 800 pixels for InLoc. DKM uses its native resolution of 880$\times$660 pixels on Aachen Day-Night v1.1, while on InLoc it follows the 800-pixel setting for semi-dense methods. RoMa consistently uses 672$\times$672 pixels on both Aachen Day-Night v1.1 and InLoc.

\begin{table*}[htb]
	\centering
    \caption{Results of visual localization on Aachen Day-Night Dataset (Outdoor) and InLoc Dataset (Indoor). All methods are categorized into three groups: sparse, semi-dense, and dense.}
    	\label{tab:localization}
			\begin{tabular}{l l c c c c}
			\toprule
			\multirow{3}{*}{Category} & \multirow{3}{*}{Method} & \multicolumn{2}{c}{Aachen Day-Night} & \multicolumn{2}{c}{InLoc} \\
            \cmidrule(l){3-6}
            & & \multicolumn{4}{c}{(0.25m, 2\textdegree) / (0.5m, 5\textdegree) / (1.0m, 10\textdegree)} \\
            \cmidrule(l){3-6}
			& & Day & Night & DUC1 & DUC2 \\
			\midrule			
			\multirow{9}{*}{Sparse}
			& SuperGlue & 87.9 / 95.0 / 97.9 & 84.7 / 92.9 / 99.0 & 44.9 / 66.2 / 78.8 & 46.6 / 74.0 / 77.1 \\
            & ParaFormer & 86.7 / 92.4 / 97.2 & 81.6 / 90.8 / 98.0 & -- / -- / -- & -- / -- / -- \\
            & SAM & 89.2 / 94.2 / 97.8 & 84.0 / 92.6 / 98.5 & -- / -- / -- & -- / -- / -- \\
            & LightGlue & 88.0 / 93.8 / 97.5 & 84.7 / 91.8 / 99.0 & 44.0 / 64.1 / 75.8 & 42.7 / 67.9 / 73.3 \\
            & SGMNet & 86.5 / 93.2 / 97.2 & 82.7 / 91.8 / 99.0 & 39.9 / 56.6 / 70.2 & 39.7 / 59.5 / 65.6 \\
            & ResMatch & 86.8 / 93.7 / 97.2 & 81.6 / 91.8 / 98.0 & 42.9 / 61.6 / 73.7 & 38.2 / 62.6 / 69.5 \\
			& DiffGlue & 88.3 / 95.3 / 97.8 & 85.7 / 93.9 / 99.0 & 46.5 / 67.2 / 78.3 & 49.6 / 71.8 / 76.3 \\
			& SemaGlue & 88.6 / 95.1 / 97.8 & 86.7 / 91.9 / 99.0 & 47.5 / 68.2 / 80.3 & 47.3 / 73.3 / 75.6 \\
            & MambaGlue & 89.0 / 95.3 / 98.7 & 86.7 / 93.9 / 100.0 & -- / -- / -- & -- / -- / -- \\
			\midrule
			
			\multirow{13}{*}{Semi-Dense}
			& LoFTR & 88.7 / 95.6 / 99.0 & 78.5 / 90.6 / 99.0 & 47.5 / 72.2 / 84.8 & 54.2 / 74.8 / 85.5 \\
            & RCM & 89.7 / 96.0 / 98.7 & 72.8 / 91.6 / 99.0 & -- / -- / -- & -- / -- / -- \\
			& TopicFM & 90.2 / 95.9 / 98.9 & 77.5 / 91.1 / 99.5 & 52.0 / 74.7 / 87.4 & 53.4 / 74.8 / 83.2 \\
			& ASpanFormer & 89.4 / 95.6 / 99.0 & 77.5 / 91.6 / 99.5 & 51.5 / 73.7 / 86.0 & 55.0 / 74.0 / 81.7 \\
			& PATS & 89.6 / 95.8 / 99.3 & 73.8 / 92.1 / 99.5 & 55.6 / 71.2 / 81.0 & 58.8 / 80.9 / 85.5 \\
			& ElofTR & 89.6 / 96.2 / 99.0 & 77.0 / 91.1 / 99.5 & 52.0 / 74.7 / 86.9 & 58.0 / 80.9 / 89.3 \\
            & JamMa & 87.7 / 95.1 / 98.4 & 73.3 / 91.6 / 99.0 & 47.5 / 67.2 / 78.3 & 35.9 / 53.4 / 69.5 \\
			& CoMatch & 89.4 / 95.8 / 99.0 & 78.5 / 91.6 / 99.5 & 54.5 / 75.3 / 86.9 & 59.5 / 84.7 / 87.8 \\
			& EDM & 89.1 / 96.2 / 98.8 & 77.0 / 92.1 / 99.5 & 51.5 / 72.7 / 85.9 & 59.5 / 82.4 / 88.5 \\
			& AffineFormer & 89.9 / 96.2 / 98.9 & 79.1 / 91.1 / 99.5& 56.1 / 74.7 / 86.9 & 55.0 / 79.4 / 87.0 \\
			& CasP & 89.2 / 96.1 / 98.9 & 78.0 / 91.6 / 99.5 & 52.0 / 77.3 / 86.4 & 55.0 / 80.2 / 84.0 \\
			& CasMTR & 90.4 / 96.2 / 99.3 & 78.5 / 91.6 / 99.5 & 53.5 / 76.8 / 85.4 & 51.9 / 70.2 / 83.2 \\
			& IMAmatch & 90.5 / 96.4 / 99.5 & 80.1 / 91.3 / 99.6 & 53.6 / 74.8 / 84.1 & 56.5 / 78.6 / 83.2 \\						
			\midrule
			
			\multirow{2}{*}{Dense}
			& DKM & 88.1 / 95.3 / 98.5 & 72.3 / 91.1 / 97.9 & 50.5 / 73.7 / 84.8 & 53.4 / 72.5 / 74.0 \\
			
			& RoMa & 88.1 / 95.6 / 98.4 & 71.7 / 90.1 / 97.9 & 55.6 / 77.3 / 88.4 & 59.5 / 80.9 / 83.2 \\
			\bottomrule
		\end{tabular}
\end{table*}

\textbf{Results.}
Visual localization results reported in Table \ref{tab:localization} reveal that the three matching paradigms exhibit distinct strengths in different scenes.
For the indoor InLoc dataset, semi-dense methods demonstrate superior performance. Their ability to model dense feature correspondences effectively handles weak textures and repetitive structures. At the commonly reported threshold of \{$1.0\,\text{m}, 10^\circ$\}, the majority of semi-dense methods exceed 83\% success rate on DUC2. Sparse methods, constrained by keypoints detector, generally fall below 80\% under the same metric. Dense methods, despite generating full-image matches, do not consistently surpass semi-dense methods. For the outdoor Aachen Day-Night Dataset, recent sparse methods such as SemaGlue achieve consistent high performance in both daytime and nighttime scenes, highlighting the gains brought by enriching feature representations or refining attention mechanisms within the established sparse matching paradigm. Semi-dense methods maintain high accuracy in both daytime and nighttime conditions. Although dense methods provide richer representations, they are more susceptible to low-light noise and exhibit a significant performance drop in nighttime conditions. Overall, in visual localization experiments, semi-dense methods deliver the best performance across indoor and outdoor environments.

\section{Open Problems and Future Trends}
\label{Trends}
As cross-view feature matching continues to mature as a core capability in computer vision, its role is gradually shifting from a standalone matching problem toward a foundational component of intelligent systems. Despite substantial progress,  cross-view feature matching remains fundamentally constrained by unresolved issues in representation, estimation, and reasoning. Some major open problems are as follows.
\begin{enumerate}
    \item \textbf{Representation gap between geometric precision and semantic invariance.}
    Features optimized for precise spatial localization tend to be sensitive to appearance variations, while semantically robust features often sacrifice spatial discriminability. Current methods rely on a single embedding space insufficient for encoding both fine-grained geometry and high-level semantics. Developing hierarchically structured representations that decouple and coordinate multi-level visual cues is a promising direction.

    \item \textbf{Ambiguity and uncertainty in correspondence estimation.}
    Most existing methods model confidence or matchability to assess correspondence reliability, yet still produce largely deterministic predictions with a single best match. This is insufficient under repetitive textures, occlusions, or weak visual cues where multiple plausible matches may exist. Recent works have started to explicitly model uncertainty, for example by predicting probabilistic correspondence distributions such as in RoMa v2. However, a unified probabilistic formulation that supports multi-hypothesis matching and jointly captures matchability and uncertainty remains an open challenge.

    \item \textbf{Lack of explicit correspondence-level reasoning.}
    Most existing matchers enforce one-to-one assignment implicitly through dual-softmax or Sinkhorn normalization, but lack explicit correspondence-level reasoning during inference. While some methods incorporate epipolar geometry or cycle consistency as supervision signals during training, these constraints are rarely enforced at test time. Integrating such geometric and structural priors directly into the inference process would improve interpretability and reliability.

    \item \textbf{Limited generalization of correspondence models.}
    Models trained on datasets like MegaDepth often fail to generalize to unseen domains due to the lack of correspondence-specific inductive biases, such as geometric equivariance or appearance invariance. Designing architectures and training objectives that explicitly enforce these properties, including geometry-aware representations and transformation-consistent learning, is essential for real-world deployment.

    \item \textbf{Incomplete modeling of non-rigid and dynamic correspondence.}
    Most benchmarks and methods assume static scenes with rigid transformations, which does not reflect real-world applications involving articulated motion, occlusion, or temporal dynamics. Treating correspondence as a continuous spatiotemporal transformation problem with deformation priors and cross-frame consistency constraints remains an underexplored direction.

    \item \textbf{Lack of foundation models pretrained for geometric correspondence.}
    Existing vision foundation models are mainly pretrained for semantic understanding, segmentation, or generation, rather than cross-view correspondence learning. Although they provide robust semantic and structural priors, they lack explicit modeling of geometric transformations, spatial alignment, and fine-grained feature association required for image matching. However, the interaction between these models and specialized matching architectures, such as SuperGlue, LoFTR, or RoMa, remains largely unexplored. Developing pretraining paradigms tailored to correspondence, for example based on relative pose estimation, cycle consistency, or epipolar constraints, and bridging task-specific matchers with general 3D foundation models, represents an important future direction.

    \item \textbf{Computational bottlenecks in global correspondence modeling.}
    Global operations such as dense correlation volumes or full attention mechanisms introduce quadratic complexity, limiting scalability to high-resolution or multi-view matching. Designing scalable operators that preserve long-range dependency modeling without sacrificing global consistency or fine-grained accuracy is a pressing research need.
\end{enumerate}

\section{Conclusion}
\label{Conclusion}
This survey has presented a comprehensive and structured review of recent advances in cross-view feature matching, a fundamental problem in computer vision that has witnessed rapid progress with the advent of deep learning and VFMs. We revisited the evolution from traditional handcrafted feature matching pipelines to learning-based correspondence frameworks, highlighting the paradigm shift toward data-driven, end-to-end, and increasingly generalizable solutions. To address the fragmentation of existing research, we proposed a structured and hierarchical taxonomy that organizes current methods from the perspectives of feature representation, matching paradigms, multi-type feature integration, VFMs driven methods, and training strategies. This unified framework provides a coherent view of the design space and facilitates systematic comparison and deeper understanding of diverse approaches. Furthermore, we complemented this taxonomy with a unified benchmarking and empirical evaluation of representative methods under consistent protocols, offering fair and comprehensive insights into their performance and generalization across different scenarios. We hope that this survey provides a clear and structured foundation for understanding recent advances in cross-view feature matching and serves as a useful reference for future research toward more unified, efficient, and generalizable correspondence systems.

\ifCLASSOPTIONcaptionsoff
  \newpage
\fi



\bibliographystyle{IEEEtran}
\bibliography{egbib}

\end{document}